# SUMBT+LaRL: Effective Multi-Domain End-to-End Neural Task-Oriented Dialog System

**HWARAN LEE**[1], **SEOKHWAN JO**[2], **HYUNGJUN KIM**[2], **SANGKEUN JUNG**[3], **AND TAE-YOON KIM**[2]
[1]NAVER Green Factory, Seongnam-si, Gyeonggi 13561, South Korea
[2]SK Telecom, Jung-gu, Seoul 04539, South Korea
[3]Department of Computer Science and Engineering, Chungnam National University, Daejeon 34134, South Korea

Corresponding author: Tae-Yoon Kim (tae.y.kim@sk.com)

This work was supported in part by SK Telecom. This work was also supported by Institute of Information & communications Technology Planning & Evaluation (IITP) grant funded by the Korea government (MSIT)(No.2020-0-01441, Artificial Intelligence Convergence Research Center (Chungnam National University)).

**ABSTRACT** The recent advent of neural approaches for developing each dialog component in task-oriented dialog systems has remarkably improved, yet optimizing the overall system performance remains a challenge. Besides, previous research on modeling complicated multi-domain goal-oriented dialogs in end-to-end fashion has been limited. In this paper, we present an effective multi-domain end-to-end trainable neural dialog system *SUMBT+LaRL* that incorporates two previous strong models and facilitates them to be fully differentiable. Specifically, the SUMBT+ estimates user-acts as well as dialog belief states, and the LaRL models latent system action spaces and generates responses given the estimated contexts. We emphasize that the training framework of three steps significantly and stably increase dialog success rates: separately pretraining the SUMBT+ and LaRL, fine-tuning the entire system, and then reinforcement learning of dialog policy. We also introduce new reward criteria of reinforcement learning for dialog policy training. Then, we discuss experimental results depending on the reward criteria and different dialog evaluation methods. Consequently, our model achieved the new state-of-the-art success rate of 85.4% on corpus-based evaluation, and a comparable success rate of 81.40% on simulator-based evaluation provided by the DSTC8 challenge. To our best knowledge, our work is the first comprehensive study of a modularized E2E multi-domain dialog system that learning from each component to the entire dialog policy for task success.

## I. INTRODUCTION

Multi-domain task-oriented dialog (ToD) system aims for users to achieve multiple goals in different domains such as finding attractions and booking restaurants. Developing such a system typically requires the following dialog components to construct a pipeline as illustrated in Fig.1: natural language understanding (NLU) to extract user's intents and slot-values [1]–[4], dialog state tracking (DST) to update belief states [5]–[9], querying database (DB), dialog policy (POL) to decide the system's next action [10]–[14], and natural language generation (NLG) to generate system responses [15], [16]. Although recent advances in neural approaches in the natural language domain have greatly improved the performance of individual dialog components, errors in each component are accumulated in the pipelined system, resulting in degradation of overall performance. Therefore, designing an effective model architecture and stable optimization methods in an end-to-end fashion is still challenging.

Recently, several end-to-end (E2E) neural dialog systems have proposed [17]–[25]. First, all dialog components in the pipeline are fully differentiable and optimized in end-to-end fashion in [17], [18], but they showed success only on a simple single dialog domain. Sequence-to-sequence approaches directly generate system responses given user utterance inputs, but they have limitations that querying the external database is unavailable [26], [27], and system actions are not interpretable [19], [20]. Moreover, a few previous researchers have investigated dialog policy optimization by reinforcement learning in E2E systems [18], [28], [29], but not explored on complicated multi-domain ToD.

The associate editor coordinating the review of this manuscript and approving it for publication was Juntao Fei.

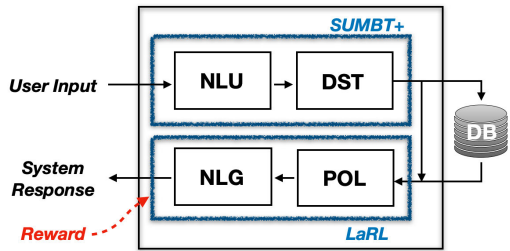

**FIGURE 1.** The conventional pipeline task-oriented dialog systems and the proposed end-to-end trainable SUMBT+LaRL.

Meanwhile, recent approaches that transfer general linguistic knowledge from large pre-trained language model, GPT-2 [30], to goal-oriented dialog have shown remarkable improvements [22], [25], [31]. They employed the GPT-2 backbone as it is, and fine-tuned the model to auto-regressively generate dialog states, system actions, and responses in a sequence. Although leveraging the rich knowledge allows the models to generate more natural and appropriate responses, reinforcement learning on transformer-based architectures has been reported as unstable [32], and learning dialog policy on those models has not been explored yet.

In this paper, we present an end-to-end trainable neural dialog system and an effective training framework with reinforcement learning for multi-domain task-completion tasks, name as *SUMBT+LaRL*. The model consists of two previously proposed components: (i) an extended version of SUMBT [5] for a word-level dialog state tracker and (ii) LaRL [33] for a word-level policy model. In addition to SUMBT that updates belief states by employing the slot-utterance matching mechanism, SUMBT+ estimates the probabilities of domains and user-intents from the user utterance, which play critical roles in the entire system performance. Then given the estimations by SUMBT+, the LaRL models categorical latent system action spaces without system action labels and generates system responses. More than other robust DST modules [6]–[9], SUMBT+ models the probability distribution of dialog states as dialog progresses thus making the entire system to be fully differentiable and enabling error back-propagation through dialog history.

In our training framework, we emphasize the importance of *separately pre-train* SUMBT+ and LaRL and *fine-tune the entire model* in an end-to-end fashion. Then, the trained dialog policy is further optimized by *reinforcement learning* using REINFORCE algorithm to succeed in dialog tasks. In the LaRL module, the policy gradients at latent actions decouple the discourse-level decision making and language generation by the decoder, enabling stable and effective reinforcement learning. Although most reinforcement training has shown increased task success rates, they often result in poor BLEU score and perplexity and tends to make hasty decisions to complete dialogs. Nevertheless, our E2E training framework shows more robust to the trade-off between the success rates and BLEU scores. We further introduce new success criteria in which the system has to respond to more requestable slots and calculate the match performance using the belief state estimated by SUMBT+.

We demonstrated the efficacy of the proposed system on MultiWOZ2.0, which is implemented in ConvLab platform for user simulator-based evaluations. Our extensive experimental results on both corpus and simulator-based evaluation show the effectiveness of the proposed pretraining and end-to-end fine-tuning framework as well as reinforcement learning of the latent action policy. From the results and qualitative analysis of simulated dialog examples, we also present the discrepancy problem of corpus-based and automatic evaluations, the limitations of corpus-based reinforcement learning, and the need for advanced reward design and success criteria to train dialog action policy. Our model achieved the new state-of-the-art success rate in the end-to-end corpus-based evaluation on the MultiWOZ2.0, as well as comparative performance to the challenge winner of the 8th dialog system technology challenge (DSTC8) challenge [3] in the simulator-based evaluation.

In summary, the main contributions of this paper are fourfold:

1) We propose an end-to-end optimizable neural dialog system, SUMBT+LaRL, which is compactly modularized and fully differentiable.
2) We present that the pre-training of each component followed by fine-tuning and reinforcement learning is a much effective optimizing framework for the entire end-to-end system.
3) We introduce new reward criteria for reinforcement learning as well as provide experimental analysis on a different result aspect depending on the success criteria and evaluation methods.
4) We present the new state-of-the-art success rate of 83.7% and 81.40% on the end-to-end corpus and simulator-based evaluation of MultiWOZ2.0, respectively.

In Sec.2, we briefly review end-to-end multi-domain task-completion dialog systems and the DSTC8 Challenge. In Sec.3, we explain the detailed architecture of the proposed SUMBT+LaRL and training procedures. Related work is described in Sec.4 and experimental results are presented in Sec.5.

## II. END-TO-END MULTI-DOMAIN TASK-COMPLETION DIALOG SYSTEMS

In this section, we overview the end-to-end multi-domain task-completion task in the DSTC8 challenge Track 1.

### A. MultiWOZ CORPUS
The goal of the challenge is to develop an end-to-end goal-oriented dialog system on MultiWOZ2.0 [34] corpus. The MultiWOZ2.0 dataset is a multi-domain task-oriented conversational corpus that helping a tourist such as booking a restaurant or finding an attraction. More specifically, it contains seven domains: *Attraction, Hospital, Hotel, Police, Restaurant, Taxi,* and *Train*. There is 10,438 annotated dialogs (8,438 train, 1,000 valid, and 1,000 test) which compose of 3,406 single-domain and 7,032 multi-domain dialogs. In multi-domain dialogs, 1 to 3 domains appear in each multi-domain dialog session.

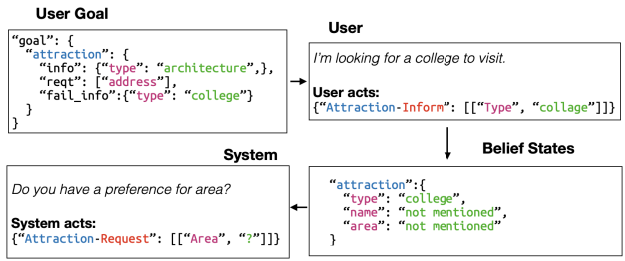

**FIGURE 2.** An example of a user goal, a user utterance, belief states, and a system utterance in a MultiWOZ dialog.

Dialog states and system dialog acts are fully annotated at each turn as described in Fig.2. User dialog acts are additionally provided in the DSTC8 challenge [3]. More specifically, dialog states refer to information shared between system and user as the dialog progresses (*i.e.*, "attraction":{"type":"college"}). Dialog acts for system or user utterances indicate semantic frames about the turn including domains, intents, slots, and slot values (*i.e.*, "Attraction-Inform": [["Type", "college"]], "Attraction-Request": [["address", "?"]]). The set of intents and slots are different for users and systems, and there are separate lists for informable slots and requestable slots. The domains, intents, slots, and slot values are pre-defined in the ontology.

### B. USER GOAL

User goals define the constraints of what users want that should be informed to systems and the information of what users have to request during the conversations. For example, the inform slot-value pairs (*i.e.*, {"type": "architecture"}) and the request slots (*i.e.*, ["entrance fee"]) are listed for each domain as user goals. Furthermore, wrong information is also stated as "fail info" to make the case that the system cannot find suitable items from the database.

### C. ConvLab AND USER SIMULATOR

DSTC8 offers an open-source platform, ConvLab[1] which allows participants to develop their systems and enables system-wise evaluation. ConvLab also provides several rule-based or neural-based dialog components of NLU, DST, POL, NLG, and end-to-end neural models. Moreover, it provides a user-simulator for automatic evaluations and a crowd-sourcing platform, Amazon Mechanical Turk, for human evaluations.

[1]https://github.com/ConvLab/ConvLab

The user-simulator consists of an RNN-based NLU model called MILU (Multi-Intent Language Understanding, [7]), a rule-based DST, an agenda-based POL, and a retrieval-based NLG. At the start of a dialog session, a user goal is randomly generated or selected among the set of pre-defined user goals. For a given user goal, the agenda-based policy [35] stacks the agenda where each entry corresponds to a pending intention, which is the user aims to achieve. The stack-update process follows complex rules. The user simulator communicates with models through natural languages by the NLU and NLG components.

### D. EVALUATION METRIC

For the automatic evaluation, dialog systems converse with the user-simulator. At the end of a dialog session, the dialog system is evaluated for how well it responded to the user's requests (*i.e.*, *Inform*) and whether the responses were consistent with the information spoken by the user (*i.e.*, *Match*). For the *Inform* category, the precision, recall, and F1 are calculated for the slots provided by the system. For the *Match* category, the system achieved *Booking Success* if the booked information matches the user's information described in the goal. Finally, when booking succeeds and inform recall is one, the dialog is *Success*.

Since the user-simulator is not perfect, the final ranking is determined by human evaluations. Crowd-workers randomly converse with submitted systems in natural languages and mark the *Success* of the user's goal as well as *natural language understanding* and *response appropriateness* on a scale of 1 to 5.

## III. SUMBT+LaRL

In this section, we describe our end-to-end dialog system. First, we introduce the overall framework and explain the details of each model architecture followed by the training procedures.

### A. OVERALL SYSTEM ACHITECTURE

As described in Fig.3, our end-to-end dialog system mainly consists of two neural sub-modules: the word-level dialog state tracker (SUMBT+) and the word-level policy model (LaRL).

There is a four-step procedure for the system to respond to the user. (i) The SUMBT+, which is an extended model of Slot-Utterance Matching Belief Tracker (SUMBT) [5], extracts domains and user intents of the current turn, and tracks belief states, simultaneously. (ii) Then, the system queries the domain-specific database (*i.e.*, restaurant, or movie) using the estimated current domain and dialog states. (iii) The Latent Action Reinforcement Learning (LaRL) module [33] implicitly models the action spaces without any system-action supervision from the utterance semantic vector, and information of extracted domains, user intents, belief states, and database query results. Then it generates the delexicalized system response. (iv) Finally, the delexicalized tokens are lexicalized using the DB query results.

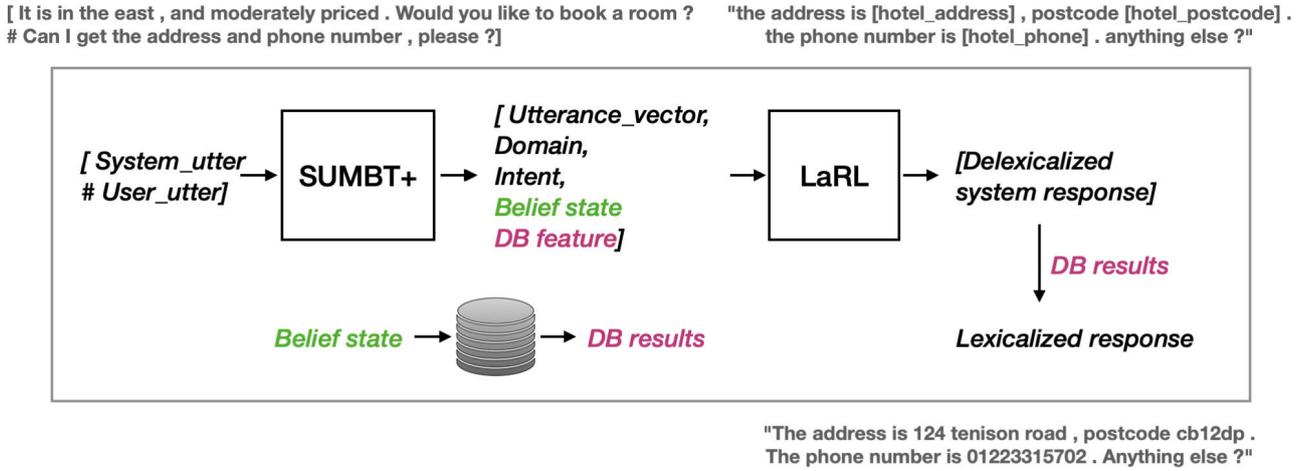

**FIGURE 3.** The overall architecture of end-to-end trainable SUMBT+LaRL.

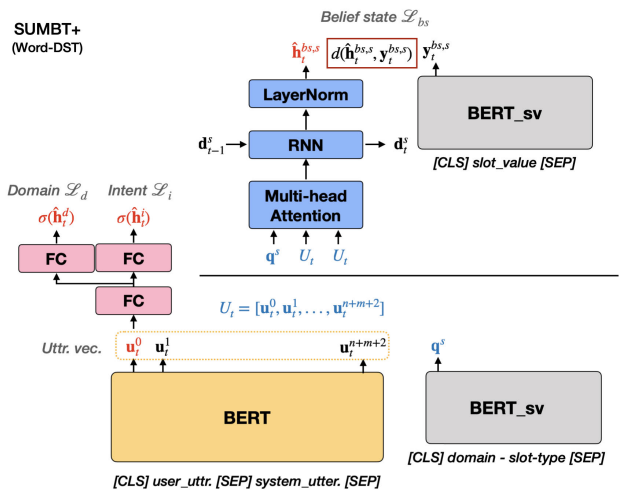

**FIGURE 4.** The architecture of extended Slot-Utterance Matching Belief Tracker (SUMBT+) that classifies domains and user actions at the dialog turn *t* and estimates belief states along the dialog.

Sec.III-B and Sec.III-C describe the model architectures of SUMBT+ and LaRL, respectively. Sec.III-D and Sec.III-E describe the end-to-end training and the reinforcement learning of the proposed system, respectively.

### B. SUMBT+: NATURAL LANGUAGE UNDERSTANDING AND DIALOG STATE TRACKING

The proposed SUMBT+ estimates domain and user intent as well as belief state at the dialog turn $t$ in multi-task learning fashion, as shown in Fig.4. Formally, for the pair of system and user utterances, $x_t^{sys} = (w_{1,t}^{sys}, \ldots, w_{n,t}^{sys})$ and $x_t^{usr} = (w_{1,t}^{usr}, \ldots, w_{m,t}^{usr})$, the model predicts domain labels $\mathbf{y}_t^d \in [0, 1]^{n_d}$, user-intent labels $\mathbf{y}_t^i \in [0, 1]^{n_i}$, and slot-values $\mathbf{y}_t^{bs,s}$ for each domain-slot-types $s$ which is one-hot vector of dimension $|\mathcal{C}_s|$, where $n_d$ and $n_i$ are number of domains and user-intents, and $\mathcal{C}_s$ is a set of the candidate slot-values of slot-type $s$ which is defined in the ontology. Note that the transition of dialog domain can occur at one dialog turn (*i.e.*, from restaurant reservation to seeking an attraction), and a user utterance can be a mixture of multiple intents (*i.e.*, 'Inform-Address' and 'Request-Name'). Therefore, they are multi-label binary classification tasks, whereas we assume only one slot-value exists for each domain-slot-type.

First, we employ the pre-trained BERT to encode each word $w$ into a contextual semantic word vector $\mathbf{u}$, and the encoded utterances are represented in the following matrix representation:

$$U_t = \text{BERT}\left(\left[x_t^{sys}, x_t^{usr}\right]\right). \quad (1)$$

We considered $\mathbf{u}_t^0$ vector as an utterance vector that represents the entire semantics of the sentence pair.

The utterance vector is then fed into the two fully connected (FC) layers to predict domain and intent labels. The network parameters are trained to minimize following the cross-entropy loss:

$$\mathcal{L}_d(\theta) = -\sum_{t=1}^{T}\sum_{d=0}^{n_d} \mathbf{y}_t^d \log \mathbf{h}_t^d + (1 - \mathbf{y}_t^d) \log(1 - \mathbf{h}_t^d),$$

$$\mathcal{L}_i(\theta) = -\sum_{t=1}^{T}\sum_{i=0}^{n_i} \mathbf{y}_t^i \log \mathbf{h}_t^i + (1 - \mathbf{y}_t^i) \log(1 - \mathbf{h}_t^i), \quad (2)$$

where $\mathbf{h}_t^d = \sigma(\hat{\mathbf{h}}_t^d)$, $\mathbf{h}_t^i = \sigma(\hat{\mathbf{h}}_t^i)$. $\sigma$ is a sigmoid function, and $\hat{\mathbf{h}}_t^d$ and $\hat{\mathbf{h}}_t^i$ are the domain and act FC layer output vectors, respectively.

For belief state tracking, the model literally encodes the words of domain-slot-types $s$ and slot-values $v_t$ at turn $t$ likewise system and user utterances. Another pre-trained BERT, denoted as $\text{BERT}_{sv}$, is fixed during the training, and it encodes their word sequences $\mathbf{x}^s$ and $\mathbf{x}_t^v$ into contextual semantic vectors $\mathbf{q}^s$ and $\mathbf{y}_t^v$, respectively.

$$\mathbf{q}^s = \text{BERT}_{\text{sv}}(\mathbf{x}^s),$$
$$\mathbf{y}_t^v = \text{BERT}_{\text{sv}}(\mathbf{x}_t^v). \quad (3)$$

We use the output vectors corresponding to the classification embedding token, [CLS], as the semantic vectors.

The model uses multi-head attention [36] to find the relevant information corresponding to the domain-slot-type from the utterances. Considering the encoded vector of the domain-slot-type $\mathbf{q}^s$ as a query, the model matches it to the contextual semantic vectors $\mathbf{u}$ at each word position, and then the attention scores are calculated. The multi-head attention maps a query matrix $Q$, a key matrix $K$, and a value matrix $V$ with different linear $h$ projections, and then the scaled dot-product attention is performed on those matrices. The attended context vector $\mathbf{h}_t^s$ between the slot $s$ and the utterances at $t$ is

$$\tilde{\mathbf{h}}_t^s = \text{MultiHead}(Q, K, V), \quad (4)$$

where $Q$ is $Q^s$, and $K$ and $V$ are $U_t$. A recurrent neural network (RNN) with a layer normalization layer inputs the attended vector and the previous RNN output vector $\mathbf{d}_{t-1}^s$ together, and outputs $\mathbf{d}_t^s$ and the predicted semantic vector $\hat{\mathbf{h}}_t^{bs,s}$. Note the RNN is shared across different domains and slot-types.

The model is trained to minimize the distance between outputs $\hat{\mathbf{h}}_t^{bs,s}$ and target slot-value's semantic vectors $\mathbf{y}_t^{bs,s}$ under a certain distance metric. The probability distribution of a slot-value $v_t$ is calculated as

$$p\left(v_t | \mathbf{x}_{\leq t}^{sys}, \mathbf{x}_{\leq t}^{usr}, s\right) = \frac{\exp\left(-d(\hat{\mathbf{h}}_t^{bs,s}, \mathbf{y}_t^{bs,v})\right)}{\sum_{v' \in \mathcal{C}_s} \exp\left(-d(\hat{\mathbf{h}}_t^{bs,s}, \mathbf{y}_t^{bs,v'})\right)}, \quad (5)$$

where $d$ is a distance metric such as Euclidean distance or negative cosine distance. This discriminative classifier is similar to the metric learning method proposed in [37], but the distance metric is measured in the fixed space that the pretrained BERT represents rather than in a trainable space.

The network is trained to minimize the log likelihood for all dialog turns $t$ and slot-types $s \in \mathcal{D}$ as following:

$$\mathcal{L}_{bs}(\theta) = -\sum_{t=1}^{T} \sum_{s \in \mathcal{D}} \log p(v_t | \mathbf{x}_{\leq t}^{sys}, \mathbf{x}_{\leq t}^{usr}, s). \quad (6)$$

By training all domain-slot-types together, the model can learn general relations between slot-types and slot-values, leading to performance improvements.

Finally, the model is trained to minimize all above losses:

$$\mathcal{L}_{DST}(\theta) = \mathcal{L}_{bs}(\theta) + \lambda_d \mathcal{L}_d(\theta) + \lambda_i \mathcal{L}_a(\theta), \quad (7)$$

where $\lambda_d$ and $\lambda_i$ are hyper-parameters.

### C. LATENT ACTION SPACE AND RESPONSE GENERATION

To model latent system action spaces and generate responses, we employed the LaRL model proposed in [33]. As described in Fig.5, the context encoder estimates $p(\mathbf{z}|\mathbf{c})$, the probability of latent variable $\mathbf{z}$ given context vectors $\mathbf{c}$. In our system, the context vector concatenates the utterance vector $\mathbf{u}^0$, the domain probability vector $\mathbf{h}^d$, the action probability vector $\mathbf{h}^i$, the belief state probability vector $\mathbf{h}^{bs}$, and the database query result vector $\mathbf{h}^{db}$. Note that the belief state probability in $\mathbf{h}^{bs}$ represents the estimated probability that the slot-value is *not* 'none' or 'not mentioned'. Also, the database query results, such as the number of retrieved results and domain activation, are one-hot encoded into the vector.

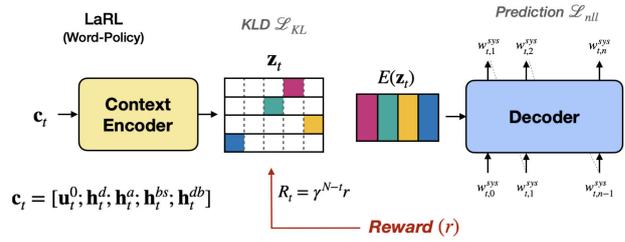

**FIGURE 5.** The Latent Action Reinforcement Learning (LaRL) model with Categorical latent variables.

The latent action variables can be modeled as continuous and discrete variables. Instead of continuous variables with Gaussian prior distribution, in this paper, we assume them as $M$ independent $K$-way categorical random variables with Uniform prior distribution, resulting in $K^M$ combinations for unique action spaces. The probability of $Z_m$ is estimated by the context encoder $F$ and a softmax function as follow:

$$p(Z_m|\mathbf{c}) = \text{softmax}(\pi_m(F(\mathbf{c}))),$$
$$\mathbf{z}_m \sim p(Z_m|\mathbf{c}). \quad (8)$$

For each categorical random variable $Z_m$, there is a $D$-dimensional vector space $\mathbf{E}_m \in \mathbb{R}^{K \times D}$. Therefore, the sampled discrete latent variables $\mathbf{z}_m$ are then mapped into the vector space using their embeddings $\mathbf{E}_m(\mathbf{z}_m) \in \mathbb{R}^D$.

Finally a RNN decoder such as LSTM or GRU autoregressively generates a response given the latent embedding vectors:

$$p(\mathbf{x}^{sys}|\mathbf{z}) = p(\mathbf{x}^{sys}|[\mathbf{E}_1(\mathbf{z}_1); \ldots; \mathbf{E}_M(\mathbf{z}_M)]; \theta_d). \quad (9)$$

The summation of latent embedding vectors are fed into the initial state of decoder. In addition, the decoder selectively integrates latent information into decoding process by using attention mechanism:

$$\alpha_{mi} = \text{softmax}(\mathbf{h}_i^T W_a \mathbf{E}_m(\mathbf{z}_m)),$$
$$\mathbf{c}_i = \sum_{m=1}^{M} \alpha_{mi} \mathbf{E}_m(\mathbf{z}_m),$$
$$\tilde{\mathbf{h}}_i = \tanh W_s([\mathbf{h}_i; \mathbf{c}_i]),$$
$$p(w_i|\mathbf{h}_i, \mathbf{c}_i) = \text{softmax}(W_o \tilde{\mathbf{h}}_i), \quad (10)$$

where $\mathbf{h}_i$ is the RNN output at $i$-th step, and $W_a$, $W_s$, $W_o$ are learnable parameters. Thus the selective fusion enables different information fusion at each generation step.

The network is optimized to maximize the evidence lower bound (ELBO) via stochastic variational inference [38] in auto-encoding manner:

$$\mathcal{L}_{full}(\theta) = \mathbb{E}_{q(\mathbf{z}|\mathbf{x},\mathbf{c})}[p(\mathbf{x}|\mathbf{z})] - D_{KL}\left[q(\mathbf{z}|\mathbf{x},\mathbf{c})||p(\mathbf{z}|\mathbf{c})\right], \quad (11)$$

where $q(z|x,c;\phi)$ is the approximate posterior distribution parameterized by neural networks, and $D_{KL}$ is KL-divergence. The ELBO is maximized by gradient ascent via a reparameterized trick. We use the Gumbel-Softmax [39] to backpropagate through categorical latent variables.

Nevertheless, ELBO training usually yields discrepancy problems between latent variables sampled from posterior and prior distributions, which results in poor generation results. In order to avoid this exposure-bias problem, the authors in [33] proposed a simplified ELBO by assuming the posterior and prior network are the same (*i.e.*, $q(\mathbf{z}|\mathbf{x},\mathbf{c};\phi) = p(\mathbf{z}|\mathbf{c};\theta)$), and adding the additional regularization term $\beta D_{KL}\left[q(\mathbf{z}|\mathbf{c})||p(\mathbf{z})\right]$. The regularized term prevents overfitting and the posterior being apart from the prior. $\beta$ is a hyperparameter between 0 and 1, and the prior distribution $p(\mathbf{z})$ is Uniform distribution for categorical variables. Finally, the loss function of the policy model is the negative simplified ELBO:

$$\mathcal{L}_{POL}(\theta) = -\mathbb{E}_{p(\mathbf{z}|\mathbf{c})}[p(\mathbf{x}|\mathbf{z})] + \beta D_{KL}\left[p(\mathbf{z}|\mathbf{c})||p(\mathbf{z})\right]. \quad (12)$$

### D. PRETRAINING AND END-TO-END FINE-TUNING

Since the proposed neural dialog system is fully differentiable, the entire system can be optimized in end-to-end fashion by minimizing the total loss:

$$\mathcal{L} = \mathcal{L}_{DST} + \lambda \mathcal{L}_{POL}. \quad (13)$$

However, the total loss function is a combination of different types of losses that have different ranges of scales, so we expect that optimizing the whole system from scratch usually causes some criteria to be under-fitted. Therefore we separately *pretrain* SUMBT+ and LaRL models while freezing BERT encoder, and *fine-tune* the entire system with the loss Eq.13.

### E. REINFORCEMENT LEARNING OF DIALOG POLICY

After end-to-end training, the dialog system is further optimized by policy gradient reinforcement learning [40], [41] using the REINFORCE algorithm [10] to succeed task-dependent goals. For the system response at turn $t$, a turn-level reward $r_t$ is given, and the expected discounted return through $T$-length dialog is calculated as $J(\theta) = \mathbb{E}[\sum_0^T \gamma^t r_t]$ with the discounting factor $\gamma \in [0, 1]$. In order to reduce the variance of the policy gradient, the reward uses a baseline function $b$ is as follow:

$$R_t = \sum_{k=0}^{T-t} \gamma^k (r_{t+k} - b). \quad (14)$$

**TABLE 1.** The proposed reward criteria.

| Reward Level | Requestable Slots Belonging to Reward Criteria |
| --- | --- |
| 1 (baseline) | 'phone', 'address', 'postcode', 'reference', 'trainid' |
| 2 | + 'leaveAt', 'arriveBy', 'type' |
| 3 | + 'price', 'duration', 'entrance' |

Then, the reward signal is backward to the latent action space by REINFORCE algorithm as follow:

$$\nabla_\theta J(\theta) = \mathbb{E}_\theta \left[\sum_{t=0}^T R_t \nabla_\theta \log p_\theta(\mathbf{z}|\mathbf{c}_t)\right]. \quad (15)$$

Note that the policy gradient at latent actions decouples the discourse-level decision making from language generation of the decoder, and encourages more stable reinforcement learning than policy gradient at the decoder outputs.

When the dialog system succeeds the user's task goal at the end of the dialog, it achieves a reward $r_T$. That is, the model is required to offer appropriate information to the user's goal conditions, and provide all requested information by the user (*i.e.*, 'what is the address of the restaurant?'). However, the existing success criterion in corpus-based evaluation [34] includes only a part of the requestable slot set such as 'phone' and 'address', leading to the discrepancy to a real evaluation environment. Therefore, we introduce more difficult and realistic reward criteria by incrementally including requestable slots such as 'leave at' and 'price' as described in Table 1. As the number of slots criteria increases, the rewards become sparse and reinforcement learning is hard to train the dialog policy.

Reward calculation and reinforcement learning procedures are as follow:

1) For each RL episode, randomly sample a dialog from the training set.
2) At each pair of the previous system and the user turns, obtain SUMBT+ outputs, sample the latent system actions, and generate a delexicalized system response.
3) With the estimated belief states by SUMBT+, query the database and lexicalize the system response.
4) According to the reward level, compute reward based on the generated responses.
5) Compute policy gradient using Eq.15 and update the parameters.

Note that we use the estimated belief state rather than the original belief state during the training in order to encourage the model to learn the accurate rewards.

### IV. RELATED WORK

In this section, we review related work in terms of word-level dialog state tracking, word-level action policy and response generation, and end-to-end neural task-oriented dialog systems.

*A. DIALOG STATE TRACKING*

Recent neural dialog state tracking methods are categorized into two categories: classification among slot-value candidates [5], [7], [42], and generation slot-values [6], [9]. While the classification-based methods can obtain the probability distribution for the candidates, the generation-based methods are robust against out-of-vocabulary words by finding text spans. However, the latter methods have disadvantages that it requires slot-value normalization and it is hard to estimate the probability of generated slot-value over the candidates. Note also that SUMBT models the changes of belief states as dialog progress using neural network, not by rules, which enables error back-propagation through dialog histories.

*B. ACTION POLICY AND RESPONSE GENERATION*

To learn dialog policy and response generation given dialog contexts, the existing approaches can be carefully divided into two categories depending on explicit or implicit action spaces. The explicit approaches are typically trained to classify the system actions with label supervision [16], [21], or generate dialog acts and response simultaneously [43]. However, the annotated system action labels are noisy and biased, thus resulting in performance bottlenecks of the entire system. Implicit approaches model the latent action spaces such as Gaussian or Categorical variables and apply reinforcement learning [33] or hierarchical reinforcement learning in [44] to optimize the action policy for evaluation metrics as rewards. Nevertheless, those methods were trained and evaluated on oracle dialog contexts rather than end-to-end fashion.

*C. END-TO-END NEURAL TASK-ORIENTED DIALOG SYSTEMS*

Firstly, research in modularized end-to-end dialog systems is closely related to our work. Prior work [17] introduced a pipelined neural system consisting of five modules (Intent, Belief Tracker, Policy, Database, Generation), which are trained in supervised learning. LIDM [18] learned the underlying system actions with latent variables in a semi-supervised setting, then it was refined using RL with turn-level rewards, and it was demonstrated on a single domain dialog. Our work differs from the prior studies in the way that we focus on large-scale and more complex multi-domains, a less modularized pipeline structure, unsupervised learning for system action spaces, and reinforcement learning for dialog-level policy optimization.

Another line of research based on sequence-to-sequence models has been proposed to reduce accumulated errors. Sequicity [19] incorporates dialogue state tracking and it is jointly trained with response generation. MOSS [45] has a shared encoder and used modular supervisions likewise multi-task learning, and UniConv [24] generates dialog acts and response simultaneously. DAMD [23] aims to the one-to-many problem of dialog response, suggesting multi-action generation.

A recent line of research is employing the large-scale pre-trained language model, GPT-2 [30], and transfer the knowledge into task-oriented dialogs. They use the GPT-2 model as a pipeline without any additional module network. The researchers in [46] fine-tuned GPT-2 on the MultiWOZ dataset, but they only dealt with oracle dialog context-to-text tasks. The challenge winner of the DSTC8 task [22] input dialog history to GPT-2 and generated belief states, system actions, and responses auto-regressively. Similarly, Simple-TOD [31] and SOLOIST [25], which are parallel but independent work, generate belief states, and then database query results are fed into the network, followed by response generations. Meanwhile, the difficulty of reinforcement learning on the transformer-based architecture has not been fully resolved yet [32], and thus dialog policy optimizations on those approaches require further research. Although the GPT-2 eases the human efforts for modeling each module, the auto-regressively generating all belief states, the system acts, and responses in a sequence would be a speed bottleneck when it comes to real-world applications.

To our best knowledge, our work is the first comprehensive study of a modularized E2E multi-domain dialog system that learning from each component to the entire dialog policy for task success.

## V. EXPERIMENTAL RESULTS

In this section, we experimentally demonstrate the efficacy of end-to-end fine-tuning of the pretrained SUMBT+LaRL model. Furthermore, we show that reinforcement learning improves the dialog success rate of the model in both end-to-end corpus-based and simulator-based evaluations, indicating the necessity and effectiveness of reinforcement learning. We present the experiment results with dialog examples depending on the proposed reward levels and the sort of belief states for reward calculations. Finally, we show a new state-of-the-art dialog success rate on end-to-end corpus-based evaluation as well as our DSTC8 challenge result, which is comparative to the challenge winner.

*A. EXPERIMENTAL SETTINGS*

1) PREPROCESSING

Following the conventional delexicalization method in [34], we replaced the user and system utterances with placeholders. We additionally delexicalized several slot-values based on the user-act annotations provided by DSTC8. For the database features, we also followed the way described in [34] and concatenated them with query result-dependent binary features, such as '*parking*' or '*internet*'.

2) TRAINING DETAILS

We developed our system employing the pre-trained $BERT_{BASE}$ model,[2] and open-source repositories for SUMBT[3] and LaRL.[4] Overall, the SUMBT+ predicts 7 domains, 40 user intents, and slot-values for 38 slots in belief states. We described the hyperparameters used for our experimental results in Table 11, Appendix A. Note that in the case of reinforcement learning, we only trained the context encoder to fine-tune the parameters for probability function given the context vector, $p(\mathbf{z}|\mathbf{c})$.

---
[2] https://github.com/huggingface/transformers
[3] https://github.com/SKTBrain/SUMBT
[4] https://github.com/snakeztc/NeuralDialog-LaRL

**TABLE 2.** Belief States, User Action, and Domain Classification Results of Pretrained SUMBT+.

|  | Belief States | | Domain | | | User Intents | | |
|---|---|---|---|---|---|---|---|---|
|  | Joint | Slot | Acc. | Precision | Recall | Acc. | Precision | Recall |
| SUMBT [5] | 0.4881 | 0.9733 | - | - | - | - | - | - |
| SUMBT+ (Fix BERT) | 0.4901 | **0.9759** | 0.9696 | 0.8274 | 0.9165 | 0.9778 | 0.5843 | 0.7892 |
| SUMBT+ (Finetune BERT) | **0.4920** | 0.9750 | **0.9896** | **0.9528** | **0.9622** | **0.9865** | **0.7695** | **0.8688** |

**TABLE 3.** Corpus-based Evaluation on Pretrained LaRL.

| Posterior | PPL | Success | Inform | BLEU | F1 |
|---|---|---|---|---|---|
| Lite [33] | 4.05 | 0.574 | 0.680 | **19.1** | N/A |
| Lite | **4.02** | **0.685** | **0.776** | 18.89 | 0.510 |

### 3) EVALUATION METRICS

For the SUMBT+, we measure the joint accuracy of belief state tracking and the classification accuracies of both domain and user intents. For the LaRL and the entire system, the evaluation metrics are categorized into corpus-based evaluations and simulator-based evaluations. For corpus-based evaluations, the performance of the MultiWOZ test split is measured in the perspective of success rate, inform rate, delexicalized-BLEU, and entity F1. The Success and Inform rate are whether the model provides requested information, and the model's recommendations are suited for the user's request. Entity F1 evaluates the entity coverage accuracy. For an overall measure, a combined score are calculated as $(Inform + Success) \times 0.5 + BLEU$. For simulator-based evaluations, the model converses with the simulator provided in ConvLab for 100 sessions, then the generated dialogs are evaluated for the dialog success ratio, book rate, inform, and the number of dialogue turns.

### B. PRETRAINING SUMBT+ AND LaRL

The pretraining SUMBT+ results are shown in Table 2. Compared to SUMBT [5], SUMBT+ shows the similar belief state accuracy. Besides, although the classification accuracy of domains and user intents are high, the precision and recall are low because of the small number of positive samples. When the BERT encoder is fixed, the precision and recall are relatively less optimized than when the BERT parameters are fine-tuned. The BERT encoder will be fine-tuned in the following end-to-end fine-tuning step.

Table 3 presents the corpus evaluation results on the pretrained LaRL with the oracle labels. Compared with the results reported in [33], our model achieves a higher success rate by utilizing BERT encoder, user domain and intent labels as inputs, supplemental delexicalization, and database features.

### C. END-TO-END FINE-TUNING

End-to-end training results of our proposed model are shown in Table 4 including classification and corpus-based evaluation. In the table, we compared four end-to-end models: the combination of the pretrained models without fine-tuning (Comb. PTR), the end-to-end trained from scratch (E2E Scratch), the fine-tuned model except for the BERT encoder (E2E FT (Fix BERT)), and the end-to-end fine-tuned model (E2E FT).

First, when the two pre-trained SUMBT+ and LaRL are combined as a pipeline, the success rate degrades from 68.5% (pretrained LaRL) to 51.3% due to the accumulated errors in the pipeline. In the E2E FT model, the errors are reduced by back-propagation through the pipeline, resulting in the significant improvement of the success rate of 66.2%, as well as other classification metrics, perplexity, and BLEU scores. On the other hand, when training the entire model from scratch without pretraining procedures, it was difficult to optimize the system, resulting in a poor success rate of 31.5%. In addition, fine-tuning but fixing the BERT model significantly decreases the success rate (42.6%) and classification results, indicating that fine-tuning the BERT utterance encoder plays the key role in the E2E fine-tuning.

We also trained models without domain and user intent information to verify the impact of this information on the overall system performance. These models showed a significantly lower success rate (26.9% for the combination model, 28.3% for end-to-end trained from scratch), indicating that understanding the user's intents, especially recognizing user requests, is critical to carry on successful dialogs.

### D. REINFORCEMENT LEARNING OF DIALOG POLICY

The dialog policy of the above end-to-end fine-tuned model (E2E FT) was further trained using REINFORCE algorithm with the three reward levels in Table 1. We also experimented to compare the use of the estimated belief states and oracle belief states when querying the database and calculating rewards during the training. Note the estimated ones are used in test phases. We will analyze and discuss the evaluation results by both corpus-based and user-simulator-based evaluation methods.

**TABLE 4.** Classification and End-to-end Corpus-based Evaluation Results on SUMBT+LaRL.

|  | Belief States | | Domain | | | User Intents | | | NLG | Corpus-based Evaluation | | | |
| --- | --- | --- | --- | --- | --- | --- | --- | --- | --- | --- | --- | --- | --- |
|  | Joint | Slot | Acc. | P. | R. | Acc. | P. | R. | PPL | Success | Inform | BLEU | F1 |
| Comb. PTR w/o UA | 0.481 | 0.974 | - | - | - | - | - | - | 5.130 | 0.269 | 0.424 | 14.22 | 0.347 |
| E2E Scratch w/o UA | 0.481 | 0.974 | - | - | - | - | - | - | 5.139 | 0.283 | 0.442 | 14.22 | 0.347 |
| Comb. PTR | 0.491 | 0.976 | 0.970 | 0.827 | 0.917 | 0.978 | 0.584 | 0.789 | 4.211 | 0.513 | 0.708 | 16.25 | 0.432 |
| E2E Scratch | 0.484 | 0.975 | 0.990 | 0.954 | 0.965 | 0.987 | 0.767 | 0.876 | 4.810 | 0.359 | 0.548 | 15.48 | 0.380 |
| E2E FT (Fix BERT) | 0.488 | 0.977 | 0.959 | 0.749 | 0.903 | 0.971 | 0.411 | 0.735 | 4.112 | 0.426 | 0.716 | 16.77 | 0.430 |
| E2E FT | **0.515** | **0.979** | **0.991** | **0.957** | **0.967** | **0.991** | **0.861** | **0.904** | **3.999** | **0.662** | **0.721** | **19.36** | **0.522** |

**TABLE 5.** End-to-end Corpus-based Evaluation on REINFORCE Trained SUMBT+LaRL.

| Model | Reward Level | PPL | Success | Match | BLEU | F1 | Combined | Success 2 | Success 3 |
| --- | --- | --- | --- | --- | --- | --- | --- | --- | --- |
| Lite + RL [33] | - | 5.22 | 0.792 | 0.828 | 12.8 | N/A | 93.8 | N/A | N/A |
| Comb. PTR | - | 4.21 | 0.513 | 0.708 | 16.25 | 0.432 | 61.21 | 0.41 | 0.30 |
| E2E FT | - | **4.00** | 0.662 | 0.721 | **19.36** | **0.522** | 88.51 | 0.54 | 0.44 |
| Oracle BS | 1 | 4.08 | 0.828 | 0.912 | 18.60 | 0.504 | **105.60** | 0.70 | 0.58 |
|  | 2 | 4.06 | 0.811 | 0.890 | 18.80 | 0.508 | 103.85 | 0.67 | 0.55 |
|  | 3 | 4.08 | 0.796 | 0.856 | 18.40 | 0.494 | 101.00 | 0.66 | 0.53 |
| Estimated BS | 1 | 4.10 | **0.854** | 0.922 | 17.90 | 0.501 | 106.70 | **0.72** | **0.62** |
|  | 2 | 4.09 | 0.850 | 0.924 | 18.40 | 0.501 | **107.10** | **0.72** | 0.59 |
|  | 3 | 4.09 | 0.804 | 0.880 | 18.40 | 0.490 | 102.60 | 0.67 | 0.53 |

**TABLE 6.** Simulator-based Evaluation on REINFORCE Trained SUMBT+LaRL.

| Model | Reward Level | Turn | Inform | | | Book Rate (%) | Success (%) |
| --- | --- | --- | --- | --- | --- | --- | --- |
|  |  |  | Precision | Recall | F1 |  |  |
| Comb. PTR | - | 12.17 | 0.60 | 0.73 | 0.63 | 66.48 | 57.20 |
| E2E FT | - | 9.04 | **0.69** | 0.86 | 0.74 | 80.36 | 77.40 |
| Oracle BS | 1 | 9.18 | 0.64 | 0.85 | 0.71 | 84.30 | 74.20 |
|  | 2 | 8.50 | 0.68 | **0.90** | **0.75** | **86.53** | **81.40** |
|  | 3 | 7.88 | 0.67 | 0.87 | 0.73 | 83.64 | 79.20 |
| Estimated BS | 1 | 10.63 | 0.60 | 0.78 | 0.65 | 77.90 | 65.60 |
|  | 2 | 8.27 | 0.63 | 0.86 | 0.70 | 81.08 | 76.60 |
|  | 3 | 8.17 | 0.68 | 0.88 | 0.74 | 83.76 | 79.00 |

### 1) END-TO-END CORPUS-BASED EVALUATION

Table 5 shows the corpus-based evaluation results, and the success rates on the proposed reward criteria are denoted as Success 2 and Success 3.

First, after RL training, the success rate is improved from 66.2% (E2E FT) to 85.4% (Estimated BS with Reward Level 1). Our models even outperform the reinforcement trained LaRL [33] that uses the oracle context inputs to generate responses. Also, note that the BLEU and perplexity of our proposed architecture show relatively less degradation after reinforcement learning than [33]. In addition, as we hypothesized before that the use of the estimated BS by SUMBT+ produce more accurate reward signal, those models show improved success and match rates than the models using the oracle BS. Regarding the reward levels, the models trained with the more difficult success criteria (higher reward level) perform the lower success rates. For the reason behind this result, we contend that the model was difficult to appropriately respond to every requested slots and receive reward signals, leading to inefficient reinforcement learning.

### 2) SIMULATOR-BASED EVALUATION

The trained models are automatically evaluated by the user-simulator and the results are presented in Table 6. First of all, our fully differentiable architecture enables the errors between the modules to be optimized in the end-to-end fashion, resulting in a significant improvement from the combined pretrained models (57.20%) by achieving success rate of 77.40%.

Furthermore, the reinforcement trained model results in the best success rate of 81.40%. Overall, we observe that 1) booking rate increases, and 2) the precision and recall of Inform decrease after the reinforcement learning which implies the models provide additional information at once to achieve rewards. In addition, 3) reinforcement learning induces the model to finish and success in shorter turns.

**TABLE 7.** Dialog examples of REINFORCE trained models with different reward levels. Slot-values in user utterances are underlined.

| | |
|---|---|
| Goal | {'attraction': {'info': {'area': 'centre', 'type': 'park'}, 'reqt': {'entrance fee': '?'}}, 'train': {'info': {'destination': 'ely', 'day': 'friday', 'departure': 'birmingham new street', 'leaveAt': '21:45'}, 'reqt': {'price': '?'}}}" |
| SUMBT+LaRL + RL w/ Oracle BS (Reward Level 1): Dialog Failed | |
| User | **Okay , are there any parks in the centre ? I am also looking for places to go in town . Maybe something in the centre .** |
| – Inferred user action | {'Attraction-Inform (BS)': [['Area', 'centre'], ['Type', 'park']], 'Inform-Inform': [['Area', ''], ['Type', '']]}, |
| System | **there are 1 park in the centre. cambridge university botanic gardens is located in the centre and cambridge university botanic gardens is 4 pounds.** |
| – Inferred system action | {'Attraction-Inform': [['Name', 'cambridge university botanic gardens'], ['Area', 'centre'], ['Name', 'cambridge university botanic gardens'], ['Fee', '4 pounds'], ['Type', 'park'], ['Choice', '1'], ['Area', 'centre']]}, |
| User | **I am leaving for ely. I would like to leave after 21:45 . I will departing from birmingham new street . I ' m looking to travel on friday.** |
| – Inferred user action | {'Train-Inform (BS)': [['Day', 'friday'], ['Depart', 'birmingham new street'], ['Dest', 'ely'], ['Leave', '21:45']], 'Inform-Inform': [['Day', '21:45'], ['Depart', '21:45'], ['Dest', '21:45']]}, |
| System | **tr8272 leaves at 23:00 and arrives at 23:51. would you like me to book that for you?** |
| – Inferred system action | {'Train-Inform': [['arrive', '23:51'], ['Id', 'tr8272'], ['Leave', '23:00']], 'Train-OfferBook': [['none', 'none']]}, |
| User | <span style="color:red">**What is the price of a ticket , please ?**</span> |
| – Inferred user action | {'Request-Request': [['Ticket', '?']]} |
| System | <span style="color:red">**I have a train tr8272.**</span> |
| – Inferred system action | {'Train-Inform': [['Id', 'tr8272']]} |
| User | <span style="color:red">**What is the price for the train ?**</span> |
| – Inferred user action | {'Request-Request': [['Ticket', '?']]} |
| System | <span style="color:red">**I have a train tr8272.**</span> |
| – Inferred system action | {'Train-Inform': [['Id', 'tr8272']]} |
| User | **Thank you , goodbye.** |
| SUMBT+LaRL + RL w/ Oracle BS (Reward Level 2): Dialog Success | |
| User | **Okay , are there any parks in the centre ? I am also looking for places to go in town . Maybe something in the centre .** |
| – Inferred user action | {'Attraction-Inform (BS)': [['Area', 'centre'], ['Type', 'park']], 'Inform-Inform': [['Area', ''], ['Type', '']]}, |
| System | **there is 1 park in the centre. it is called cambridge university botanic gardens. would you like more information?** |
| – Inferred system action | {'Attraction-Inform': [['Name', 'cambridge university botanic gardens'], ['Choice', '1'], ['Area', 'centre'], ['Type', 'park']], 'general-reqmore': [['none', 'none']]}, |
| User | **Is there an entrance fee ?** |
| – Inferred user action | {'Attraction-Inform (BS)': [['Name', 'cambridge university botanic gardens']], 'Request-Request': [['Fee', '?']]}, |
| System | **cambridge university botanic gardens is free. their phone number is 01223336265.** |
| – Inferred system action | {'Attraction-Inform': [['Fee', 'free'], ['Name', 'cambridge university botanic gardens'], ['Phone', '01223336265']]}, |
| User | **Yes I would like to go to ely please . Yes , I need to leave sometimes after 21:45 . Thanks ! I also need a train departing from birmingham new street . I will leave on friday .** |
| – Inferred user action | {'Train-Inform (BS)': [['Day', 'friday'], ['Depart', 'birmingham new street'], ['Dest', 'ely'], ['Leave', '21:45']], 'Inform-Inform': [['Day', '21:45'], ['Depart', '21:45'], ['Dest', '21:45']]}, |
| System | **tr8272 leaves at 23:00 and arrives at 23:51. would you like me to book that for you?** |
| – Inferred system action | {'Train-Inform': [['Arrive', '23:51'], ['Leave', '23:00'], ['Id', 'tr8272']], 'Train-OfferBook': [['none', 'none']]}, |
| User | <span style="color:red">**I need to know the price.**</span> |
| – Inferred user action | {'Request-Request': [['Ticket', '?']]} |
| System | <span style="color:red">**the price is 23.60 pounds.**</span> |
| – Inferred system action | {'Train-Inform': [['Ticket', '23.60 pounds']]}, |
| User | **Ok , have a good day . Goodbye .** |

Next, we observe the different tendencies of evaluation results according to the corpus and simulator-based evaluations. In the user-simulator evaluation, training with the oracle belief states gains larger improvements than the estimated belief states. We interpret the reason behind this phenomenon as due to the discrepancy between corpus-based training and online simulator-based evaluation. In other words, the off-line corpus-based RL has the limitation that the environment assesses the dialog success after the dialog ends, but it does not produce turn-level rewards. Moreover, the true pairs of system and user utterances are given to the model rather than the generated system utterance at the previous turn.

Besides, lower reward levels degrade the success rates whereas they were effective in the corpus-based evaluation.

**TABLE 8.** End-to-end Corpus-based Evaluation on MultiWOZ.

| Models | Success | Match | BLEU | Combined |
|---|---|---|---|---|
| Sequicity [19] | 45.3 | 66.4 | 15.54 | 71.41 |
| HRED-TS [47] | 58.0 | 70.0 | 17.50 | 81.50 |
| Structured Fusion [21] | 58.6 | 73.8 | 16.90 | 83.10 |
| Neural pipeline GPT-2 (Challenge Winner) [22] | 62.4 | 73.0 | 16.00 | 83.70 |
| DAMD [23] | 60.4 | 76.3 | 16.60 | 84.95 |
| UniConv [24] | 62.9 | 72.6 | **19.80** | 87.55 |
| SimpleTOD [31] | 70.1 | 84.4 | 15.01 | 92.26 |
| SOLOIST [25] | 72.9 | 85.5 | 16.54 | 95.74 |
| SUMBT+LaRL (E2E Finetune) | 66.2 | 72.1 | 19.36 | 88.51 |
| SUMBT+LaRL + RL w/ Oracle BS | 82.8 | 91.2 | 18.60 | **105.6** |
| SUMBT+LaRL + RL w/ Estimated BS | **85.4** | **92.2** | 17.90 | **106.7** |

**TABLE 9.** DST Performance Comparison with Previous Work.

| Models | Joint (%) | Slot (%) |
|---|---|---|
| SUMBT [5] | 48.81 | 97.33 |
| TRADE [6] | 48.62 | 96.22 |
| COMER [8] | 48.79 | - |
| DSTQA [7] | 51.44 | 92.74 |
| SOMDST [9] | **51.72** | - |
| SUMBT+ (FT BERT) | 49.01 | 97.59 |
| SUMBT+LaRL (E2E FT, Fix BERT) | 48.79 | 97.69 |
| SUMBT+LaRL (E2E FT) | **51.52** | **97.89** |

**TABLE 10.** DSTC8 Results on Automatic Evaluation by User-Simulator [†]: our submission applied post-processing.

| Team | E2E trainable | Turn | Inform P. | R. | F1 | BR. (%) | Success (%) |
|---|---|---|---|---|---|---|---|
| T1 |  | 7.00 | 0.92 | 0.96 | 0.93 | 93.75 | 88.80 |
| T2 |  | 6.69 | 0.83 | 0.94 | 0.87 | 96.39 | 88.60 |
| T3 |  | 6.55 | 0.71 | 0.92 | 0.78 | 94.56 | 82.20 |
| T10[†] *(ours)* | ✓ | 8.50 | 0.68 | 0.90 | 0.75 | 86.53 | 81.40 |
| T4 |  | 7.21 | 0.78 | 0.89 | 0.81 | 86.45 | 80.60 |
| T5 | ✓ | 7.59 | 0.8 | 0.89 | 0.83 | 87.02 | 79.40 |
| T6 |  | 7.90 | 0.61 | 0.73 | 0.64 | 75.71 | 58.00 |
| T7 |  | 9.78 | 0.68 | 0.77 | 0.70 | 58.63 | 56.60 |
| T10 *(ours)* | ✓ | 8.83 | 0.46 | 0.75 | 0.54 | 76.38 | 52.20 |

We contend that the models were over-fitted to the easy success criteria and they were trained to generate specific responses to attain the easy rewards that require fewer slots, thereby losing the ability to respond to other contexts. We further analyze the case with dialog examples in the next Sec.V-E.

### E. ANALYSIS ON SIMULATED DIALOGS

The dialog examples between the REINFORCE trained SUMBT+LaRL and the user simulator are presented in Table 7. Given the same user goal, we compare the case when the model trained with reward level 1 fails the task and when the model successes with reward level 2. At each turn, the user and system utterances, the inferred user actions (belief states and user intents) by the model, and the inferred system actions by the user-simulator are described. We emphasized the part of dialogs as red texts; when the user asks the price of the offered ticket, the success model gives the appropriate answer despite the wrong inferred user intents. However, the failed model provides the train id rather than the price, indicating the model is overfitted to the success criteria by reinforcement learning. Therefore, the example presents the discrepancy problem between corpus and automatic evaluations, and the limitations of off-line reinforcement learning for dialog systems. It also suggests the need for advanced reward design and success criteria.

### F. COMPARISON WITH PREVIOUS WORK

In Table 8, the end-to-end corpus-based evaluation performance of SUMBT+LaRL is compared with the previously proposed models. The SUMBT+LaRL with reinforcement learning obtained a new state-of-the-art success rate of 85.4%, a match of 92.2%, and a combined score of 106.7. Compared to the previous GPT-2 based models such as Neural pipeline GPT-2 (The Challenge Winner) [22], SOLOIST [25], and SimpleTOD [31], the performance is largely improved, encouraging and emphasizing the importance of policy optimization through reinforcement learning.

Additionally, when it comes to dialog state tracking (DST) task, the performance of SUMBT+ (Fine-tune BERT) is similar to the one of SUMBT as 49.01% and 48.81%, respectively. This indicates that the additional supervision by domain and user intent information does not enhance the performance of DST itself. However, SUMBT+LaRL (E2E FT) shows the competitive dialog state tracking performance (51.52%) with state-of-the-art methods. It means that the supervision from the next system response by backpropagation through pipeline encourages SUMBT+ to predict the belief state more accurately.

### G. DSTC8 CHALLENGE RESULTS

In the DSTC8 challenge, the submissions were evaluated by the user simulator provided in ConvLab and human evaluators through Amazon Mechanical Turk. In the automatic evaluation, an agenda-based user simulator with MILU and a template-based NLG carried on 500 dialogs with each submitted system. For human evaluation, we refer to the challenge results [48] (See Appendix C for the details).

The automatic evaluation results are listed in Table 10, and we marked each model whether it is end-to-end trainable. Except for our SUMBT+LaRL (T10) and Neural pipeline GPT-2 (T5, the challenge winner), the other submitted models mainly have a pipeline of 4 different modules (NLU-DST-Policy-NLG). First, the submission models (T1, T2, T4) consist of BERT-based NLU, and rule or template-based DST, Policy, and NLG. T3 is similar to the first group but employed the DQN model for Policy, and HDSA with templates for NLG. T6 and T7 put their efforts to enhance policy learning. Except for T5 and our model (T10), all others include handcrafted rules and templates, so end-to-end optimization is not possible.

Since our submission had a lack of several necessary post-processing methods for lexicalization and querying databases, we polished them by following the methods used in the T5. (See Appendix B for the details). As a result, our model achieves a success rate of 81.40% in automatic evaluation (denoted as $^\dagger$). Our competitive result indicates room for replacing the sophisticated and complex hand-crafted rules with the proposed end-to-end trainable SUMBT+LaRL.

## VI. CONCLUSION

In this paper, we propose an end-to-end trainable and compactly modularized dialog system, SUMBT+LaRL. The SUMBT+ estimates user-acts as well as dialog belief states, and the LaRL models latent system action spaces and generates responses given the estimated contexts. We experimentally demonstrated that the training framework in which the SUMBT+ and LaRL are separately pretrained, then the entire system is fine-tuned significantly increases dialog success rates. Besides, the end-to-end fine-tuning showed that the supervisions from the next system response also encourage to improve dialog state tracking, achieving the performance comparable to the state-of-the-art joint accuracy. We further trained the model with reinforcement learning using the reward levels and success criteria we designed. The experiment results and the qualitative analysis of simulated dialog examples present the discrepancy problem of corpus and automatic evaluations, the limitations of corpus-based reinforcement learning for dialog policy, and the need for advanced reward design and success criteria. Our model achieved the new state-of-the-art success rate of 85.4% in end-to-end corpus-based evaluation on MultiWOZ corpus, as well as comparative performance to the challenge winner of the DSTC8 Track 1 challenge by showing 81.4% in the simulator-based evaluation.

## APPENDIX A
## TRAINING DETAILS

The following Table 11 lists the hyperparameters used for our experimental results.

**TABLE 11.** Training details for each training steps.

| Pretraining SUMBT+ | |
|---|---|
| RNN | GRU (300, 1 layer) |
| Dropout | 0.5 |
| Optimizer | BertAdam |
| Learning rate | 0.0001 |
| $\lambda_d$ | 1 |
| $\lambda_a$ | 1 |
| weight decay | 0.0002 |
| **Pretraining LaRL** | |
| Context Encoder | Linear (1 layer) |
| Decoder | Attn GRU(150) |
| Decoder vocab size | 1000 |
| Categorical z | M=10, K=20 |
| Dropout | 0.5 |
| Optimizer | BertAdam |
| Learning rate | 0.001 |
| Weight decay | 0.0002 |
| $\beta$ | 0.01 |
| Generation Beam Size | 20 |
| **End-to-end finetuning** | |
| Optimizer | BertAdam |
| Learning rate | 0.0001 |
| $\lambda$ | 1 |
| **Reinforcement Learning** | |
| Optimizer | SGD(lr=0.01, grad_clip=0.5, momentum = 0, Nestrov = False) |
| $\gamma$ | 0.99 |

## APPENDIX B
## POST-PROCESSING DETAILS

In our challenge submission, there were miscellaneous but largely influential bugs in our database query part. After fix the bugs, we applied the same post-processing methods of the challenge winner team (Neural pipeline GPT-2) [22] such as substitution several specific phrases that are not recognized by the user-simulator's NLU with other synonym words (*e.g.*, 'moderately priced' to 'moderate', 'center part of town' to 'center', 'gbp' to 'pounds'). The automatic evaluation results of our submission and its post-processing results are shown in Table 12.

**TABLE 12.** Post-processing Details.

| Model | Turn | Inform | | | BR(%) | SR (%) |
| | | P. | R. | F1 | | |
|---|---|---|---|---|---|---|
| Challenge Winner [22] | 7.59 | 0.80 | 0.89 | 0.83 | 87.02 | 79.40 |
| Our Submission | 8.89 | 0.47 | 0.76 | 0.55 | 76.50 | 50.60 |
| Fix DB Query | 8.31 | 0.56 | 0.90 | 0.65 | 86.46 | 79.60 |
| Post-processing | 8.50 | 0.68 | 0.90 | 0.75 | 86.53 | 81.40 |

## APPENDIX C
## DSTC8 HUMAN EVALUATION RESULTS

For human evaluation, crowd-workers were given randomly selected goals and had conversations with systems. After a conversation, they scored the systems for their natural lan-guage understanding and response appropriateness on a scale of 1 to 5 as well as whether the system successes user's goal.

The human evaluation results and the final challenge ranking are listed in Table 13. T5 that employed GPT-2 was ranked the first by achieving the best success rate and the remarkably highest understanding and appropriateness scores. Although our submitted model showed a lower success rate due to the post-processing problem, it achieved the 2nd rank in the average score of appropriateness (3.82) and understanding (3.78).

**TABLE 13.** DSTC8 Results on Human Evaluation.

| Team | SR(%) | Under. | Appr. | Turns | Final Ranking |
|---|---|---|---|---|---|
| T5 | 68.32 | 4.15 | 4.29 | 19.51 | 1 |
| T1 | 65.81 | 3.54 | 3.63 | 15.48 | 2 |
| T2 | 65.09 | 3.54 | 3.84 | 13.88 | 3 |
| T3 | 64.10 | 3.55 | 3.83 | 16.91 | 4 |
| T4 | 62.91 | 3.74 | 3.82 | 14.97 | 5 |
| T10 *(ours)* | 54.90 | 3.78 | 3.82 | 14.11 | 6 |
| T6 | 43.56 | 3.55 | 3.45 | 21.82 | 7 |
| T7 | 25.77 | 2.07 | 2.26 | 16.80 | 9 |

**ACKNOWLEDGMENT**
This work was supported by SK Telecom Meta AI Team for GPU cluster supports to conduct massive experiments. This work was also supported by Institute of Information & communications Technology Planning & Evaluation (IITP) grant funded by the Korea government (MSIT)(No.2020-0-01441, Artificial Intelligence Convergence Research Center (Chungnam National University)).

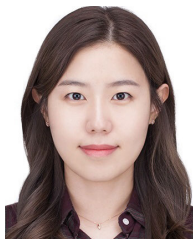
**HWARAN LEE** received the B.S. degree in mathematical science and the Ph.D. degree in electronical engineering from Korea Advanced Institute of Science and Technology (KAIST), Daejeon, Republic of Korea, in 2012 and 2018, respectively. She was a Research Scientist with SK Telecom, from 2018 to 2021. Since 2021, she has been with NAVER Corporation, Republic of Korea. Her current research interests include machine learning, deep learning, natural language processing, and dialog systems.

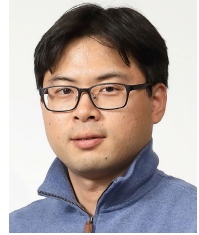
**SEOKHWAN JO** received the B.S., M.S., and Ph.D. degrees in electrical engineering from Korea Advanced Institute of Science and Technology (KAIST), Daejeon, in 2004, 2006, and 2011, respectively. From 2012 to 2018, he was a Researcher with Samsung Electronics, South Korea. Since 2019, he has been a Researcher with SK Telecom, South Korea. His research interests include signal processing, machine learning, and reinforcement learning.

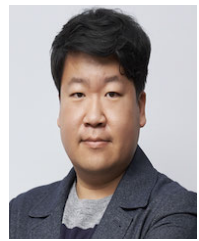
**HYUNGJUN KIM** received the B.A. degree in psychology and anthropology and the M.S. degree in computer science and engineering from Seoul National University, in 2012 and 2015, respectively. He is currently a Research Engineer with SK Telecom, Seoul, South Korea. His main research interests include goal oriented dialogue systems and lightweight deep learning models.

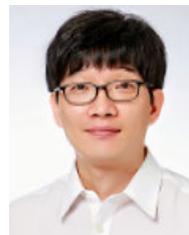
**SANGKEUN JUNG** received the B.S., M.S., and Ph.D. degrees in computer science and engineering from POSTECH, Pohang, South Korea, in 2004, 2006, and 2010, respectively. From 2010 to 2012, he was a Researcher with Samsung Electronics, Republic of Korea. From 2012 to 2014, he was a Researcher with ETRI, Daejeon, Republic of Korea. From 2014 to 2018, he was a Researcher with SK Telecom, Seoul, South Korea. Since 2018, he has been an Associative Professor with the Department of Computer Science and Engineering, Chungnam National University. His research interests include natural language processing, machine learning, and deep learning.

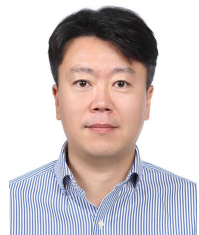
**TAE-YOON KIM** received the B.S., M.S., and Ph.D. degrees in electrical engineering from Korea University, Seoul, South Korea, in 1997, 1999, and 2006, respectively. From 2007 to 2009, he was a Visiting Researcher with the Center for Signal and Information Processing (CSIP), Georgia Institute of Technology, Atlanta, GA, USA. From 2009 to 2016, he was a Principle Research Engineer with the Samsung Research Center, Seoul. He is currently a Technical Lead with SK Telecom, Seoul, working on spoken dialogue systems. He has developed several speech recognition and language understanding engines for large scale services. His research interests include dialogue systems, speech recognition, machine learning, and statistical signal processing.

• • •